\documentclass{l4dc2024}

\title[Multi-agent transformer-accelerated RL for satisfaction of STL specifications]{Multi-agent transformer-accelerated RL for satisfaction of STL specifications}
\usepackage{times}
\usepackage{amsmath,amssymb, graphicx, multicol, array}
\usepackage[ruled]{algorithm2e}

\coltauthor{\Name{Albin {Larsson Forsberg}} \Email{albinfor@kth.se, albin.larsson.forsberg@ericsson.com}\\
 \Name{Alexandros Nikou} \Email{alexandros.nikou@ericsson.com} \\
 \Name{Aneta {Vulgarakis Feljan}} \Email{Aneta.Vulgarakis@ericsson.com}\\
 \addr Torshamnsgatan 21, 164 40 Kista, Sweden\\
 \Name{Jana Tumova} \Email{tumova@kth.se}\\
 \addr Lindstedtsvägen 24, 114 28 Stockholm, Sweden}

\begin{document}
\maketitle
\begin{abstract}%
One of the main challenges in multi-agent reinforcement learning is scalability as the number of agents increases. This issue is further exacerbated if the problem considered is temporally dependent. State-of-the-art solutions today mainly follow centralized training with decentralized execution paradigm in order to handle the scalability concerns. In this paper, we propose time-dependent multi-agent transformers which can solve the temporally dependent multi-agent problem efficiently with a centralized approach via the use of transformers that proficiently handle the large input. We highlight the efficacy of this method on two problems and use tools from statistics to verify the probability that the trajectories generated under the policy satisfy the task. The experiments show that our approach has superior performance against the literature baseline algorithms in both cases.
\end{abstract}

\begin{keywords}%
  Multi-agent reinforcement learning, Temporally dependent, Transformers, Signal temporal logic%
\end{keywords}

\section{Introduction}
Many tasks in real life are defined not only by a certain end goal, but also (maybe more importantly) by how they should get there. This could for example be a surveillance task where a security guard needs to visit multiple locations in order to check that they are safe. It is not important in what order they need to be visited in, as long as they get visited within the allotted time for the task. In other situations, we may require that subtasks happen in particular order, or that certain events happen as a response to requests. 
These types of tasks can be encoded using temporal logic specifications. Temporal logic specifications can unambigously capture spatial and temporal constraints for the task satisfaction; constraints on how states of the system should evolve over time. Due to their expressiveness they have been used to define many different tasks, such as in robotic control tasks \cite{kloetzer2007temporal}, path planning \cite{fainekos2009temporal, barbosa2019integrated}, and autonomous driving \cite{raman2015reactive, nyberg2021risk}. In particular, Signal Temporal Logic (STL) offers a rigorous way way to define desired behavior in terms of spatial-temporal constraints on continuous-time singnals \cite{maler2004monitoring}. 

Many of temporal logic-based planning and control provide good results but rely on certain assumptions to work. In general, the assumptions regard knowledge about the system model and/or the environment. It could also be the case that a model exists, but that it is too costly to use making it infeasible for real-time control. In contrast to these model-based approaches, using model-free approaches, such as reinforcement learning, to obtain a control policy handles these challenges implicitly. The control policy is learned by iteratively interacting with the environment and observing how it changes.






However, using reinforcement learning with spatial and temporal logic specifications does not come without challenges of its own. In \cite{forsberg2023network}, we showed how a spatially constrained task in a distributed multi-agent setting can be solved using coordination graphs and belief propagation algorithms. Our objective here is to solve a spatial-temporal task, i.e. history dependent task. However, a key assumption in traditional reinforcement learning is that it is \textit{Markovian}, i.e. independent of its history, which poses a challenge.
STL's expressiveness has already garnered attention within the reinforcement learning community as means of generating reward \cite{balakrishnan2019structured}. Temporal logics have has also been used in tandem with reinforcement learning to guarantee certain properties such as safety, reachability, liveness etc. \cite{jansen2020safe, bouton2018reinforcement, hasanbeig2020cautious, cohen2023temporal}. In \cite{aksaray2016q}, the authors used the quantitative robustness semantic from STL as a reward and found a policy by extending the state space. This way of solving the problem works in small scale, but as the complexity grows exponentially with the size of the state space, the method quickly looses performance when learning even with smaller problems. Instead of processing all states in one batch, \cite{liu2021recurrent} use a recurrent approach that processes the states sequentially. This approach is semi model dependent however and the specification horizon considered is also fairly short. In \cite{singh2023stl} the authors use a new way of approximating the robustness and obtain good results. However, they work with only a subset of specifications which has a reward that can be expressed well with an instantaneous reward. To a certain degree their approach addresses the question of satisfying the complexity in the temporal aspect. However, if we were to consider a multi-agent system we would suffer from similar issues related to complexity as more agents are considered. As a matter of fact, the complexity associated with time-dependence as the same as the complexity associated with multi-agent systems \cite{agogino2004unifying}.

In this work, we aim to address multi-agent reinforcement learning with spatial-temporal constraints expressed e.g., in a fragment of STL in an efficient way and we propose to leverage \emph{transformers}.
Transformers were initially presented in \cite{vaswani2017attention} as a means of solving natural language processing tasks. The paper showed the capabilities of the transformer and how it could be used to capture the inter-word dependencies in a sentence. The main contribution in their work was that they showed the expressive capabilities of attention, and how it could be used to completely replace recurrent models. Similar to how words are sequenced within sentences, states are sequenced in trajectories. This has been used advantageously in \cite{chen2021decision}, where they modeled the reinforcement learning problem as a sequence modeling problem. By training the transformer model to predict sequences of states, actions and a return-to-go, they showed that their model was able to generalize beyond information accessible in their training data in an offline learning setting.
Similarly within multi-agent systems, transformers have been used to favorably achieve state-of-the-art performance on different multi-agent reinforcement learning benchmarks, such as the Starcraft multi agent challenge. In \cite{cohen2021use}, a transformer-based architecture was used to solve the posthumous credit assignment task using counterfactual baselines and, \cite{hu2021updet}, the authors used their transformer-based approach called UPDeT to more efficiently represent the individual trajectories of each agent.

In \cite{wen2022multi}, the multi-agent transformer (MAT) is introduced. Our approach takes inspiration from their work, where they model the multi-agent reinforcement problem as a sequential decision problem. Instead of performing the sequential modeling with recurrent layers, they use a transformer based model \cite{vaswani2017attention}, mainly relying on attention in order to capture the inter-agent dependencies. The approach presented in their work handles a large number of agents efficiently, but would not scale well in a temporally dependent setting. 

The contributions of our work are two-fold:
\begin{enumerate}
    \item We present a novel transformer-based architecture that can capture both time-dependencies and inter-agent dependencies.
    \item We highlight the efficacy of our approach with experiments, and show the superiority of our method compared to baselines with statistical analysis of task satisfaction.
\end{enumerate}


\section{Preliminaries} 
In this section, the necessary notation and definitions for STL are presented. In this paper, we consider the fragment of STL recursively defined as: $$\varphi := \top\,|\,\psi\,|\,\neg \varphi\,|\, \varphi_1 \wedge \varphi_2\, | \, \lozenge_{[a,b]} \varphi,$$ where $a,b \in \mathbb{R}_{\geq 0}$ are time bounds, $\psi$ is a predicate of form $f(s) < 0$, evaluating to $\top$ if the inequality holds. The operators $\neg$ (negation) and $\wedge$ (and) can be combined to create other binary operators such as $\vee$ (or). $\lozenge$ represents the temporal operator eventually. A specific time instance of $s$ is denoted as $s_t \in S$. To be able to evaluate a specification the notation $(s,t)$ is introduced corresponding to the signal of consecutive states $s_{t'} \, \forall \, t'\in [t, \infty)$. Essentially meaning the signal from time $t$ and forward. This permits the boolean semantics of STL to recursively be described as:
\begin{align}
    (s,t) &\models f(s<\alpha) &\iff  &f(s_t) < \alpha, \nonumber\\
    (s,t) &\models \neg(f(s<\alpha)) &\iff  &\neg\left((s,t) \models f(s<\alpha)\right),\nonumber\\
    (s,t) &\models \varphi_1 \wedge \varphi_2 &\iff &(s,t) \models \varphi_1 \, \text{and} \, (s,t) \models \varphi_2, \nonumber\\
    (s,t) &\models \varphi_1 \vee \varphi_2 &\iff &(s,t) \models \varphi_1 \, \text{or} \, (s,t) \models \varphi_2, \nonumber\\
    (s,t) &\models \lozenge_{[a,b]}\varphi&\iff&\exists t' \in [t+a,t+b] \, \text{ s.t. } (s,t') \models \varphi. 
\end{align}

Moreover, a quantitative measurement called robustness, $\rho$, can be calculated for each STL specification \cite{donze2010robust}. Robustness is defined as
\begin{align}
    &\rho(s,f(s)<0,t) &\,=\, &-f(s_t), \nonumber\\
    &\rho(s,\neg (f(s)<0),t) &\,=\,  &-\rho(s,(f(s)<0),t),\nonumber\\
    &\rho(s, \varphi_1\wedge \varphi_2 ,t) &\,=\,& \min (\rho(s,\varphi_1 ,t),\rho(s,\varphi_2,t)),\nonumber\\
    &\rho(s, \varphi_1\vee \varphi_2 ,t) &\,=\,& \max (\rho(s,\varphi_1 ,t),\rho(s,\varphi_2,t)),\nonumber\\
    &\rho(s,\lozenge{[a,b]}\varphi ,t) &\,=\, &\max_{t'{\in[t+a,t+b]}}\rho(s,\varphi,t').
\end{align} A positive robustness value corresponds to a trajectory that satisfies the provided specification. Moreover, the magnitude shows how well the fit is where a higher value means the specification is satisfied with a higher margin. 

Robustness extends the notion of satisfiability beyond the binary question if it is satisfied or not and adds the ability to quantify it. This is important in order to rank two policies that both satisfy the provided specification. It can also be seen as a way to permit the policy that least violates the specification as well, e.g. if the two policies both fail. This way of reasoning over trajectories can be used to a great extent as part of a reward function as it can give partial credit to a policy that comes close to fulfilling the goal, despite not fully satisfying the goal, thus allowing for a structured way of making a sparse reward environment more dense.

\section{Problem Setting} 
The problem we consider in this paper is modeled as a partially observable Markov game $\mathcal{M}$, given as a tuple $(\mathcal{N}, S, A, R, O, P, \gamma)$; $\mathcal{N}$ is the total number of agents in $\mathcal{M}$. The agents are indexed with $I = \{1,\dots, i, \dots, \mathcal{N}\}$. Agents $1$ through $i$ are referenced as $1:i$. The set of global states is $S$, and the joint action set of all agents is $A$. The set of actions for agent $i$ is $A_i$. $R$ denotes the global reward function. $O: S \to O_i $ is the observation function which provides local observation for each agent, and $P: S\times A \to S$ denotes the transition probability function. Finally, $\gamma$ is the discount factor.


Let $\tau$ represent the trajectory of the multi-agent system from the full episode of length $T$, $(s_0, ..., s_T)$. This trajectory is generated under the joint policy of all agents $\pi$. $T$ is the fixed episode length. We also use the notation $\tau_\pi$ to denote the full trajectory generated by using policy $\pi$. The partial trajectory up until time $t$ is given as $s_{0:t}$. The reward for agent $i$ is given as $r^i_t = \rho(s_{0:t},\varphi_i)$, where $\varphi_i$ denotes the STL specification for agent $i$, defining the task over the global states, for instance constraining the distance from agent $i$ to others. The joint specification is $\varphi = \bigwedge_{i\in I} \varphi_i$. It can be observed that the reward is now temporally dependent. The objective is to maximize the expected robustness obtained from the quantitative semantics of STL.
    \begin{equation}
        \max_\pi \mathbb{E}_{\tau \sim \pi} \rho(\tau,\varphi).
        \label{eq:objective}
    \end{equation}

Existing multi-agent reinforcement learning approaches would introduce non-stationarity issues, which motivates us to approach the problem as a temporally-dependent multi-agent reinforcement learning problem. Furthermore, much of the focus today in multi-agent reinforcement learning has followed the paradigm of centralized training with decentralized execution (CTDE). 
Agent trained with CTDE-based approaches only consider their own local observations during execution. This makes them susceptible to failure if one agent unexpectedly changes its behavior. In such case all agents would need to be retrained as the policy of each agent is implicitly dependent on the local policies of other agents. A centralized execution regiment would not suffer from this issue; instead it would be able to adapt online accordingly. Centralized execution, however, is not always preferable as it leads to an exponential explosion in computational complexity. 

In this work, we propose time-dependent multi-agent transformers (TD-MAT) to allow for efficient computation of a control policy for the temporally-dependent multi-agent reinforcement learning problem in a centralized training with centralized execution (CTCE) regime.


\section{Time-dependent Multi-agent Transformer (TD-MAT)}


\begin{figure}[t!]
\centering
\includegraphics[scale=0.3]{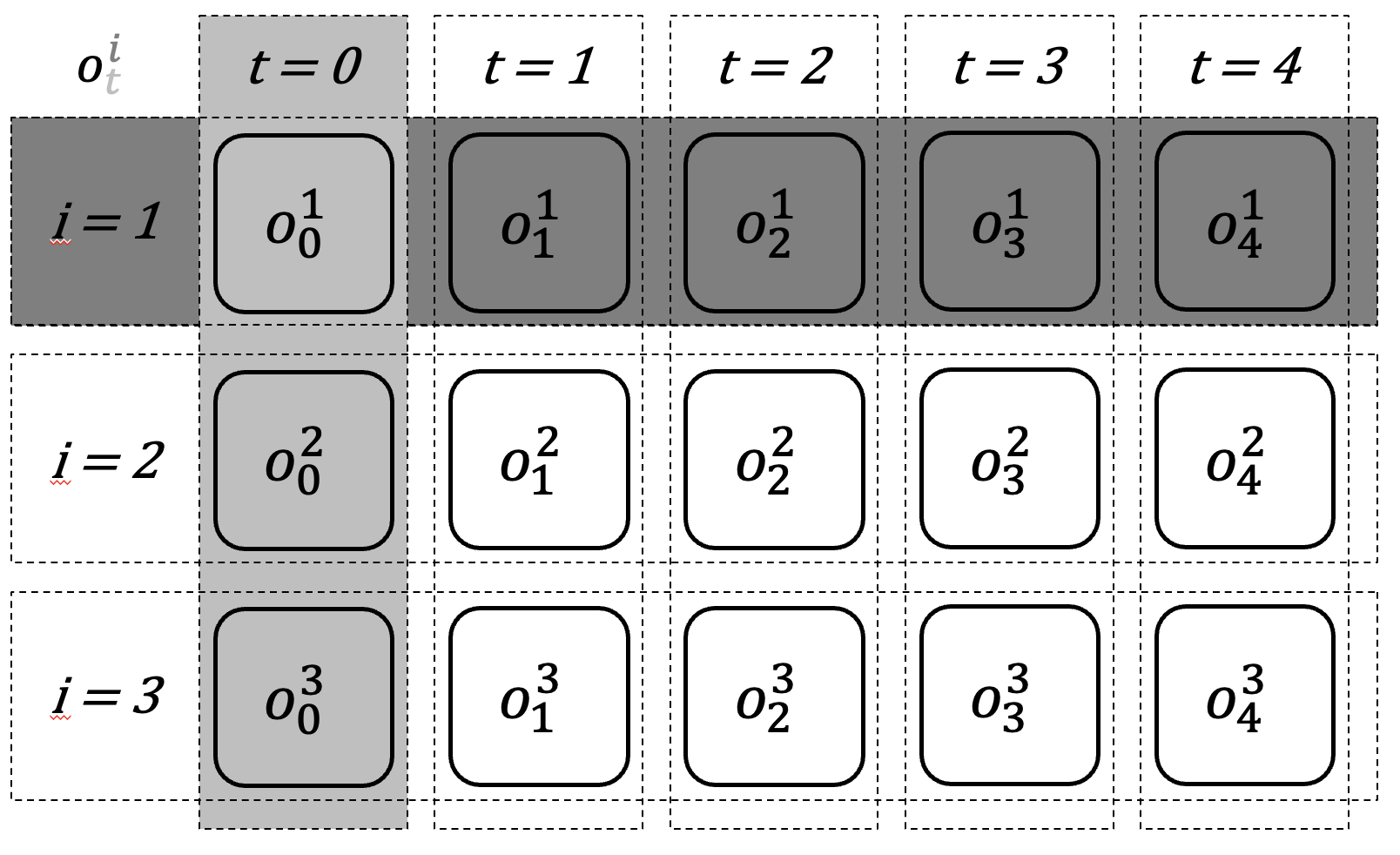}
\caption{An overview of how the encoding procedure of observations, $o$, is done. $i$ corresponds to the agent index and $t$ the timestep. They are color coded to show the time encoding axis along the vertical axis (lighter grey) and the agent encoding along the horizontal axis (darker grey).}
\label{fig:encoding}
\end{figure}

\subsection{Transformers}
Before presenting our solution, let us review basic principles of transformers. Transformers are neural networks structured to conveniently solve sequence modeling problems without relying on recurrent layers. A standard transformer consists of two parts: an encoder and a decoder. The purpose of the encoder is to, given and input $\mathbf{o}$, provide a representation $\mathbf{\hat{o}}$ that captures the interdependencies within the input sequence.  In our setting, the input is the history of the trajectory up to the current point.  This representation is then used within the decoder to provide a context for the output sequence generation process, e.g., a sequence of control inputs for RL. Both the decoder and encoder rely on the notion of attention, which is a measure of how closely related two entries in the sequence are to one another. The attention value is calculated using a query, key, and value triple, $$\text{Attention}(Q,K,V) = \text{softmax}\left(\frac{QK^T}{\sqrt{d_k}}\right)V.$$

This calculation is done in the attention-layers of a transformer. The strength of the transformer comes from the fact that each observation can be passed through the encoder in parallel, as opposed to sequentially as with GRUs, greatly speeding up inference times. In \cite{wen2022multi}, the multi-agent transformer (MAT) is introduced. MAT treats the multi-agent reinforcement learning problem as a sequential decision making problem. They do this by leveraging multi-agent advantage decomposition \cite{kuba2021settling}, which says that given an ordering of the agents, it is possible to recover the global advantage function by sequentially evaluating a local advantage function for each agent. This is done as follows: $A^{{1:n}}_\pi\left(\mathbf{o},\mathbf{a}^{{1:n}}\right)=\sum_{i=1}^n A_\pi^{i}\left(\mathbf{o},\mathbf{a}^{{1:i-1}},a^{i}\right)$. A key assumption for this theorem to hold is that the global observation is available at each evaluation step.

\subsection{Method Overview}

We use the transformer architecture to allow us to process the large-dimensional data caused by the choice of using a CTCE approach. The key idea with our approach is that input is not processed as one monolithic data input, but is instead pre-processed with positional information before it is fed to the model. This allows us to feed each input as bite-sized chunks and keep the number of model parameters low. In particular, we use a multi-variate positional encoding based on an indexation of the agents and the time at which the observation was made. Each observation (for an agent at a given time) is given as $o_t^i$, where $i$ corresponds to the agent index and t the relevant time. Figure \ref{fig:encoding} shows how such an encoding is done. Each observation gets a positional along each axes, one being the agent axis and the other being the time axis. Many different multi-variate encodings exist, but have mostly been used in for example image processing where the data is naturally distributed along multiple axes \cite{li2021learnable}. We apply a similar encoding scheme to our model where the variables we encode are the time of observation and index of the agent making the observation. By doing this encoding, we can include information from several agents as well as from several timesteps all while maintaining the same network size while processing all inputs/observation simultaneously.

\subsection{Network Structure}
\begin{figure}
\centering
\includegraphics[scale=0.25]{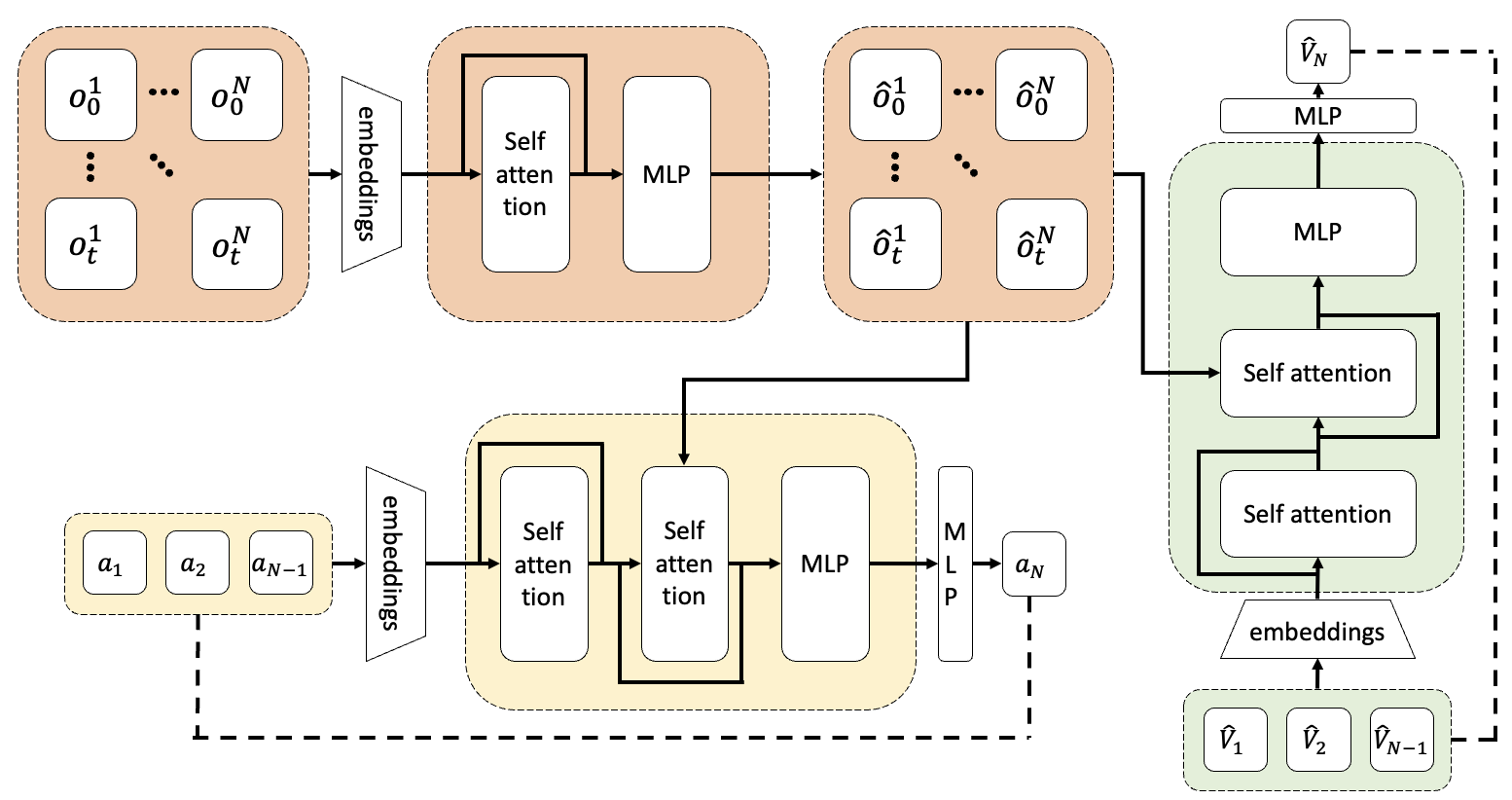}
\caption{Architectural overview of the network structure. It consists of three components (encoder, value function approximator, decoder) corresponding to the three different colors.}
\label{fig:transformer-structure}
\end{figure}

Unlike a standard transformer structure, our proposed network consists of three major components, the encoder, decoder, and the novel value function approximator. A graphical overview of the network is given in Figure \ref{fig:transformer-structure}. The three different components have been color coded accordingly. The purpose of the three components are described accordingly.

\begin{enumerate}
    \item Encoder: The encoder takes the embedded input and passes it through a number of encoder blocks. A key feature of the encoder network is that the size of the input is also the size of the output. What the encoder network tries to do is to learn the dependencies between agents and find an accurate representation that captures this. The new representation is then used as a query in the following two network components in order to provide the necessary context for their function.
    \item Value function approximator: In \cite{wen2022multi} the authors used a simple multi-layer perceptron where they mapped each observation to a value function in a one-to-one fashion. This approach would not work in our approach since if we were to do the same, we would get an excessive amount of predicted value functions. We only desire the same amount of predicted value functions as we have number of agents. Instead of a multi-layer perceptron, we use a decoder step of a transformer and generate the value functions as a sequence. We call this approximated joint value $V$. We parameterize the encoder and value network approximator jointly with $\phi$.
    \item Decoder: The purpose of the decoder is to generate a sequence of actions that is the same length as the number of agents. After this sequence has been generated, the selected actions are executed in the environment and the environment continues with a new timestep. The observation representation that is generated from the encoder is used in a masked-attention block as a query to provide the necessary context for the decoder. The decoder is parameterized with $\theta$.
\end{enumerate}

\subsection{Training}
The encoder and value network approximator are trained using a joint loss function. Note that we use several rollouts per update step. The joint latent representation from the encoder is used as query for the self-attention block of the value-network approximator. Using this context we obtain a $V_i$ for each agent. For the backward pass we calculate the loss accordingly by using the empirical Bellman error:

\begin{equation}
L_{\text{Enc\&V}}(\phi) = \frac{1}{\mathcal{N}}\sum_{i=1}^{\mathcal{N}}\sum_{t=0}^{T-1}\left[R(o_{0:t},a_{0:t})+\gamma V^i_\phi(\mathbf{\hat{o}}_{0:t+1})-V^i_\phi(\mathbf{\hat{o}}_{0:t}) \right]^2.
\label{eq:loss_enc}
\end{equation}

The values $V$ are different from other approaches in that it uses the joint observation to predict $V$ for each agent individually. One forward pass is done per timestep in order to not let the encoder have access to future observations when it creates the joint latent representation (as would be the case during rollouts). The loss for the decoder and the method to obtainthe loss is very similar as in \cite{wen2022multi}, with some important differences. The loss is given by a PPO-loss, but the policy quotient is calculated using joint observations from the beginning of the episode:

\begin{align}
L_\text{Dec}(\theta) &= -\frac{1}{\mathcal{N}}\sum_{i=1}^\mathcal{N}\sum_{t=0}^{T-1} \min\left(r^i_{0:t}(\theta)\hat{A}_{0:t},\text{clip}(r^i_{0:t}(\theta),1\pm\epsilon)\hat{A}_{0:t} \right), \label{eq:loss_dec}
\end{align}
where $r^i_{0:t}(\theta) = \frac{\pi_\theta^{i}(a^i_{t}|\mathbf{\hat{o}}_{0:t},\mathbf{a}^{1:i-1}_t)}{\pi_{\theta_\text{old}}^i(a^i_{t}|\mathbf{\hat{o}}_{0:t},\mathbf{a}^{1:i-1}_t)}$. The value for $\hat{A}_{0:t}$ is obtained using generalized advantage estimation, \cite{schulman2015high}, with the robust estimator $\hat{V}_{0:t} = \frac{1}{\mathcal{N}}\sum_{i = 1}^\mathcal{N}V^i_\phi(\mathbf{\hat{o}}_{0:t}).$

\begin{algorithm}[t!]
\scriptsize
\SetAlgoLined
\DontPrintSemicolon
\caption{TD-MAT\label{al:TD-MAT}}
\SetKwInOut{Input}{Input}
\SetKw{Parameters}{Parameters:}
\SetKw{Return}{return}
\SetKwFor{Parallel}{parallel (}{) $\lbrace$}{$\rbrace$}
\Input{\texttt{Markov game} $\mathcal{M}$, \texttt{Specification} $\varphi$}
\KwResult{\texttt{A policy} $\pi(\cdot;\phi, \theta)$}
\Parameters{\texttt{Number of iterations} $N$, \texttt{Parallel rollouts} $D$, \texttt{Learning rate} $\alpha$}
\BlankLine
\texttt{Buffer}.init()\;

$\pi \longleftarrow $ policy.init()\;

\For(\tcp*[f]{Update policy $N$ times}){$i \leftarrow 1$ \KwTo $N$}{
  \texttt{Buffer}.reset()\;
  
  \Parallel(\tcp*[f]{Do $D$ rollouts}){$D$}{
    $s \longleftarrow $ env.reset()\;
    
    \For{$t \leftarrow 0$ \KwTo $T$} {
        $\mathbf{o} \longleftarrow O(s)$\; 
        \tcp*{Make observations from state}
        $\mathbf{\Bar{o}} \longleftarrow$ pos\_enc($\mathbf{o}$)\;
        \tcp*{Add positional encoding}
        $\mathbf{a} \longleftarrow \pi(\mathbf{\Bar{o}};\phi, \theta)$\;
        \tcp*{Get actions from policy}
        $s', \mathbf{r} \longleftarrow $ env.step($\mathbf{a}$)\;
        \tcp*{Step environment forward}
        \texttt{Buffer}.insert($s,a,r,s'$)
        \tcp*{Add experience to buffer}
        $s \longleftarrow s'$
    }
  }
    $\phi \longleftarrow \phi + \alpha \nabla L_{\text{Enc\&V}}$\;
    \tcp*{Update with loss as in eq. \ref{eq:loss_enc}}
    $\theta \longleftarrow \theta + \alpha \nabla L_{\text{Dec}}$\;
    \tcp*{Update with loss as in eq. \ref{eq:loss_dec}}
  }
  \Return{$\pi(\cdot;\phi, \theta)$}
\end{algorithm}

\section{Using TD-MAT}
TD-MAT is used in an end-to-end training pipeline with the input consisting of a Markov Game (with unknown state transition probabilities), and an STL specification. In short, the Markov game defines the environment in which we act and the specification defines the task which the agents should aim to satisfy. Given these two inputs, the agents learn a behavior policy with the TD-MAT algorithm. When the user-defined number of training episodes has been reached the obtained policy is returned. One limiting factor of TD-MAT is that the frequency at which the agents send their observations to the central decision and the latency of this process, constrains in which the environment the agent can act. The trained decision model needs the local observation from all agents in order to take an action for them. A full algorithmic overview is shown in Algorithm \ref{al:TD-MAT}.

\section{Experiments}
To show the efficacy of our method, we train our model in a modified MPE environment \cite{lowe2017multi} where we have changed the reward function to be the robustness of the provided STL specification, in order to solve the optimziation problem presented in Equation \ref{eq:objective}. The state space is continuous with a size of $18$ per agent, making the joint observation space in our experiment $54$. The actions for all agents are discrete: "UP", "DOWN", "LEFT", "RIGHT", "NOTHING". The episode length during training is 25. The model will be trained across 10 million steps resulting in a total training across 400000 episodes. The training is done with 128 rollouts in parallel where each policy improvement iteration happens after these rollouts finish.

The $\mathcal{N}$ agents in the environment need to each satisfy a task $\varphi_i \forall i \in I$, where the task is provided as an STL specification. The environmental dynamics $s_{t+1} \sim P(s_t,a_t)$, are discrete and assumed to be unknown. We set the discount factor such that $\gamma = 0.99$.

We will now define the problems that will be explored. For simplicity of notation, the predicate function $\psi_i,j$ denotes the predicate $||p_i-l_j||_2 < d$. $p_i$ is the position of agent $i$ and $l_j$ is the position of landmark $j$. $d$ is the distance the agent should have to the landmark. A positive value indicates that the agent is within this desired distance. In our experiments we have disabled the collision between the agents so that they only need to focus on the task at hand. In our experiments we set $d=0.3$. The problems we consider are the following:

\begin{figure}[t!]
    \centering
    \includegraphics[width=\textwidth]{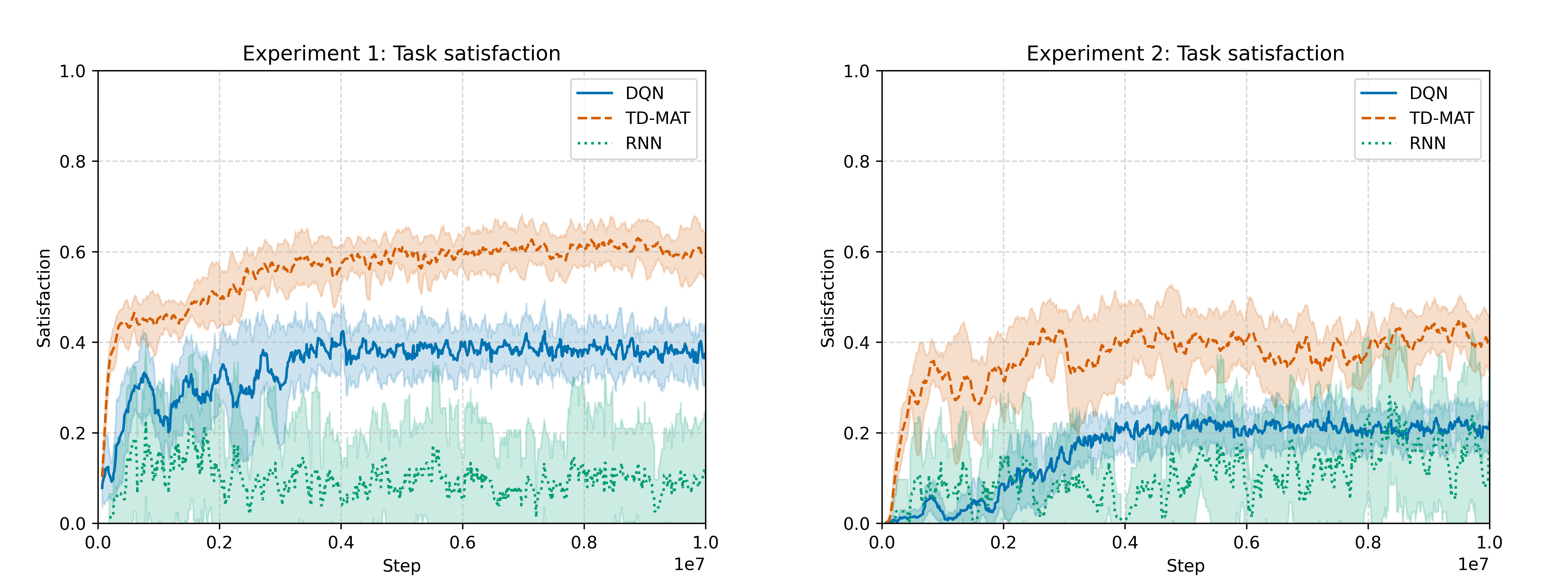}
     \caption{Average satisfaction over time for specifications for the experiments. The left graph shows results from problem 1 and the right one shows result from problem 2. Shaded areas correspond to the variance. A rolling average of $n = 5$ was used.}
     \label{fig:compare_methods}
\end{figure}

\noindent \textbf{Problem 1.1} Given the specification for agent $i$, $\varphi^1_i = \lozenge_{[0,25]}\psi_{k,k} \wedge \lozenge_{[0,25]} \psi_{k, k'}$, with $k' = (k+1) \text{ mod } \mathcal{N}$, the task is for each agent to visit two defined landmarks before the episode is over. The purpose of this task is to show that the agent is possible to learn a time-dependent task. The objective is to find a joint policy $\pi$ that can satisfy this specification. As the problem has multiple objectives, the agents need to find a policy that can differentiate between where it has been earlier in order to satsify the task.

\noindent \textbf{Problem 1.2} Given the specification for agent $i$, $\varphi^2_i = \lozenge_{[0,25]}\left(\bigwedge_{k\in I}\psi_{k,1}\right) \wedge \lozenge_{[0,25]} \psi_{i,i}$, the goal is to find a joint policy $\pi$ that allows for the trajectories generated under the policy satisfy the specification. This task means that all agents should at some time all meet at the first landmark, and also at some time visit the landmark that has a matching index to the index of the agent. Additional to being a multi-objective task, the agents need to learn how to collaborate with each other as they need to coordinate themselves to visit the same spot simultaneously. 

After the training is done we perform an analysis of the policy in order to get quantitative measures of how large the probability is that the trajectories generated under the trained policy satisfies the provided specification. We do this by using tools from statistics. The policy is evaluated in a large number of episodes, $n$, and we see if the trajectory satisfies the specification. Based on these samples we map them to a Bernoulli distribution on which we analyze the behavior. The empirical sample mean is obtained as $\hat{p} = \frac{X}{n}$ and the confidence interval in question is determined by $CI = \hat{p} \pm z_{\alpha/2}\sqrt{n\hat{p}(1-\hat{p})}$. The value $z_{\alpha/2}$ determines that with probability $1-\alpha$ our value is correct. A lower confidence leads to a tighter bound and vice-versa. For our intents and purposes we choose a confidence level of $90\%$ corresponding to a $z_{0.10/2} = 1.65$.

To the best of our knowledge, no work so far has addressed the multi-agent temporally dependent problem. We will thus highlight the capabilities of our method by comparing it to two different baselines consisting of the extended state definition and an RNN approach. The size of the input for these two methods are 1350 and 54 respectively, as opposed to 18 for our method. We ran every experiment five times. The average reward for all agents is shown in Figure \ref{fig:compare_methods} with the mean across the runs being the dashed line and the colored area indicates the magnitude of the variance. The results from problem 1 are shown in the left graph and the results from problem 2 are shown in the right graph. 

\begin{table}[t!]
    \centering
    \begin{tabular}{c|c|c|c|c|c|c|}
        \multicolumn{1}{c}{} & \multicolumn{1}{c}{} & \multicolumn{5}{c}{$P(\tau_\pi \models \varphi)$} \\ \hline
        & Model & 1 & 2 & 3 & 4 & 5\\ \hline
         &TD-MAT & $\mathbf{68.8 \pm 1.5 \%}$ & $68.1 \pm 1.5 \%$ & $58.7 \pm 1.6\%$ & $ 64.7 \pm 1,6\%$ & $52.5 \pm 1.6\%$\\
         T1 & DQN & $38.8 \pm 1.6\%$ & $36.0 \pm 1.6\%$ & $ 38.0 \pm 1.6\%$ & $38.3 \pm 1.6\%$ & $37.5 \pm 1.6\%$\\
         &RNN & $7.5 \pm 0.9\%$ & $20.0 \pm 1.3\%$ & $17.8 \pm 1.2\%$ & $9.5 \pm 1.0\%$ & $17.2 \pm 1.2\%$\\\hline
         &TD-MAT & $49.5 \pm 1.6\%$ & $46.8 \pm 1.6\%$ & $46.5 \pm 1.6\%$ & $\mathbf{51.9 \pm 1.6\%}$ & $32.9 \pm 1.5\%$\\
         T2 & DQN & $22.5 \pm 1.4\%$ & $22.4 \pm 1.4\%$ & $23.2 \pm 1.4\%$ & $21.5 \pm 1.3\%$ & $21.4 \pm 1.3\%$\\
         &RNN & $17.7 \pm 1.2\%$ & $26.3 \pm 1.4\%$ & $37.0 \pm 1.6\%$ & $36.4 \pm 1.6\%$ & $28.8 \pm 1.5\%$\\
    \end{tabular}
    \caption{A collection of how likely the obtained polices are to satisfy the given specification for trajectories generated under the policy. The confidence interval is also provided as percentage points. The numbers were obtained using $n = 2560$ and a confidence level of $90\%$. The best policies are marked in bold.}
    \label{tab:confidence}
\end{table}

We also take the obtained policies of our approach and compare it to baseline policies and evaluate how likely the obtained policies are that they satisfy the specifications. Each policy is rolled out $2560$ times and the results are seen in Table \ref{tab:confidence}. The best result we obtained from using TD-MAT for the two tasks are $68.8\%$ and $51.9\%$. This is better than the best results of $ 38.8\%$ and $37.0\%$ we got from using the baseline approaches.

\section{Discussion}
Use of the TD-MAT algorithm allows for the learning of multi-objective time-dependent behavior in a multi-agent setting. This is shown in the two experiments that we did. The first experiment was easier for the agents to satisfy, as the task satisfaction for each agents were not dependent on each other to be able to satisfy the task. In the second experiment they needed to coordinate closely. In these experiments, it is shown that TD-MAT is superior compared to the baseline approaches. In the first experiment we see that it is easier for the agent to learn a task where the input considered is of significant size. Despite addressing the problem from a centralized perspective, the decision making model manages to efficiently handle the input due to the representative power the transformer architecture has at a scale. We also observe worse performance for the more difficult task in experiment 2. This is expected as the problem is of a more complex nature. Irregardless of this, we do see that our method beats the baseline in this experiments anyway.

In Table \ref{tab:confidence}, we see that even the worst policy with TD-MAT beats the best non-TD-MAT policy in experiment one, and only two non-TD-MAT policies beat the worst TD-MAT policy. These two policies that beat our method are however varying quite a lot between different training runs. Even though it achieved a result of $37.0\%$ in one run, it also achieved a result of $17.7\%$ in another whereas the TD-MAT and Q-learning based approaches are more stable in their performance across different runs.

\section{Conclusion}
TD-MAT is a novel method that scales well with the number of timesteps considered as well as the number of agents. We have highlighted this in our experiments where we show two different tasks, where our method beats the baselines in terms of task satisfaction. The efficacy of our method is validated with statistical analysis of the probability that our method satisfies the task we have provided for it. The novelty of our method lies in the fact that we encode our input and how the transformer architecture allows for an efficient processing of full trajectory data for all agents in parallel. In future work, we will be looking into how we can expand the fragment of STL that we are working with to encode for more diverse behavior, such as safety and reach-and-hold behavior and we will also look into different ways of encoding the STL specification such that the policy can be trained on several different specifications and allow for seamless switching between behaviors. 

\clearpage 

\acks{This project is financially supported by the Swedish Foundation for Strategic Research. The research has been carried out as part of the Vinnova Competence Center for Trustworthy Edge Computing Systems and Applications at KTH Royal Institute of Technology. We also want to thank Prof. David Parker for sharing his knowledge on quantitative verification analysis.}

\bibliography{references}

\end{document}